\documentclass[conference]{IEEEtran}
\usepackage{times}

\usepackage[numbers]{natbib}
\usepackage{multicol}
\usepackage[bookmarks=true]{hyperref}
\usepackage{graphicx}

\begin{document}

\title{Green for Go, Red for No: Visual Grounding via Semantic Segmentation for VLA Navigation Policies}



\author{\authorblockN{Adrian Szvoren}
\authorblockA{Department of Computer Science\\
University College London\\
adrian.szvoren.23@ucl.ac.uk}
\and
\authorblockN{Dimitrios Kanoulas}
\authorblockA{Department of Computer Science\\
University College London\\
d.kanoulas@ucl.ac.uk}
\and
\authorblockN{Nilufer Tuptuk}
\authorblockA{Department of Security\\ and Crime Science\\
University College London\\
n.tuptuk@ucl.ac.uk}}


%

\maketitle

\begin{abstract}
Vision-language-action (VLA) models enable robot navigation from natural language and visual goals, but remain susceptible to perceptual distractions and ambiguous scene interpretations. This paper presents the first empirical evaluation of visual grounding for VLA navigation policies. We propose a real-time segmentation-based grounding method that highlights traversable areas in green and non-traversable areas in red using SegFormer. Two variants are evaluated: observation-only segmentation and joint observation-goal augmentation. Using OmniVLA on the Grand Tour dataset, we show that visual grounding reduces the mean waypoint error by $27-44\%$ at the farthest waypoint, depending on the instruction length. The benefits are greater for long instructions than for short instructions, and grounding provides little improvement for image goals. Normalized error analysis indicates that grounding primarily acts as a trajectory length regularizer, reducing the predicted path length by $30\%$ without improving per-unit-distance reasoning. Our results indicate that visual grounding offers a simple, computationally inexpensive method to improve VLA navigation without model retraining, although it cannot compensate for missing training signals in out-of-distribution instructions.
\end{abstract}

\IEEEpeerreviewmaketitle

\section{Introduction}

Vision-language-action (VLA) models have emerged as a promising development for robot navigation, enabling agents to interpret visual observations and natural language instructions to generate actions or navigation trajectories. Unlike traditional navigation systems that often rely on pre-built maps or explicit waypoint sequences, VLA models leverage large-scale pretraining on diverse datasets to generalize to novel environments and instructions without task-specific fine-tuning. Recent advances have demonstrated impressive capabilities in following free-form language goals and adapting to unseen spatial layouts \cite{hirose2025omnivla, cheng2024navila, chiang2024mobility}. However, despite these successes, VLA policies remain susceptible to perceptual distractions, ambiguous scene interpretations, and the inherent difficulty of translating high-dimensional visual input into precise navigational actions.

One promising direction for improving VLA robustness is visual grounding. Visual grounding techniques augment or preprocess visual input to make spatial relationships and traversability cues more explicit to the underlying model. In the context of vision-language models, methods have shown that overlaying images with segmentation masks, numerical labels, or action visualizations can significantly enhance reasoning and visual question answering accuracy \cite{yang2023set, nasiriany2024pivot}. This leads to the following research question: can visual grounding similarly benefit navigation VLA policies, where the agent must continuously reason about traversable space, obstacles, and goal-directed movement in real-time?

To empirically evaluate this, we instantiate two variants of our grounding pipeline: one that segments only the current visual observation, and a second that additionally augments the goal modality (appending explicit traversability instructions to language goals or segmenting image goals). We focus on three key constraints: real-time inference, generalization to unseen environments, and multimodal goal support.

The key contributions of this paper are: $(1)$ to our knowledge, the first empirical evaluation of visual grounding for a VLA navigation policy; $(2)$ a real-time segmentation-based grounding method using SegFormer~\cite{xie2021segformer}; $(3)$ two grounding variants for ablative analysis; $(4)$ quantitative results showing that grounding reduces far-waypoint error by $27–44\%$ depending on the instruction length, with greater benefits for long than short instructions, and minimal improvement for image goals; and $(5)$ identification of the ``stop'' instruction failure mode in current VLA models.

The remainder of this paper is organized as follows. Section~\ref{sec:related-work} reviews related work on VLM and VLA navigation and visual grounding. Section~\ref{sec:problem-formulation} formalizes the trajectory generation problem. Section~\ref{sec:method} describes our segmentation-based grounding method and implementation choices. Section~\ref{sec:experiments} presents the experimental setup, evaluation metrics, and quantitative results. Section~\ref{sec:limitations} discusses limitations and Section~\ref{sec:conclusion} concludes with directions for future work.

\begin{figure*}[h]
    \centering
    \includegraphics[width=1\linewidth]{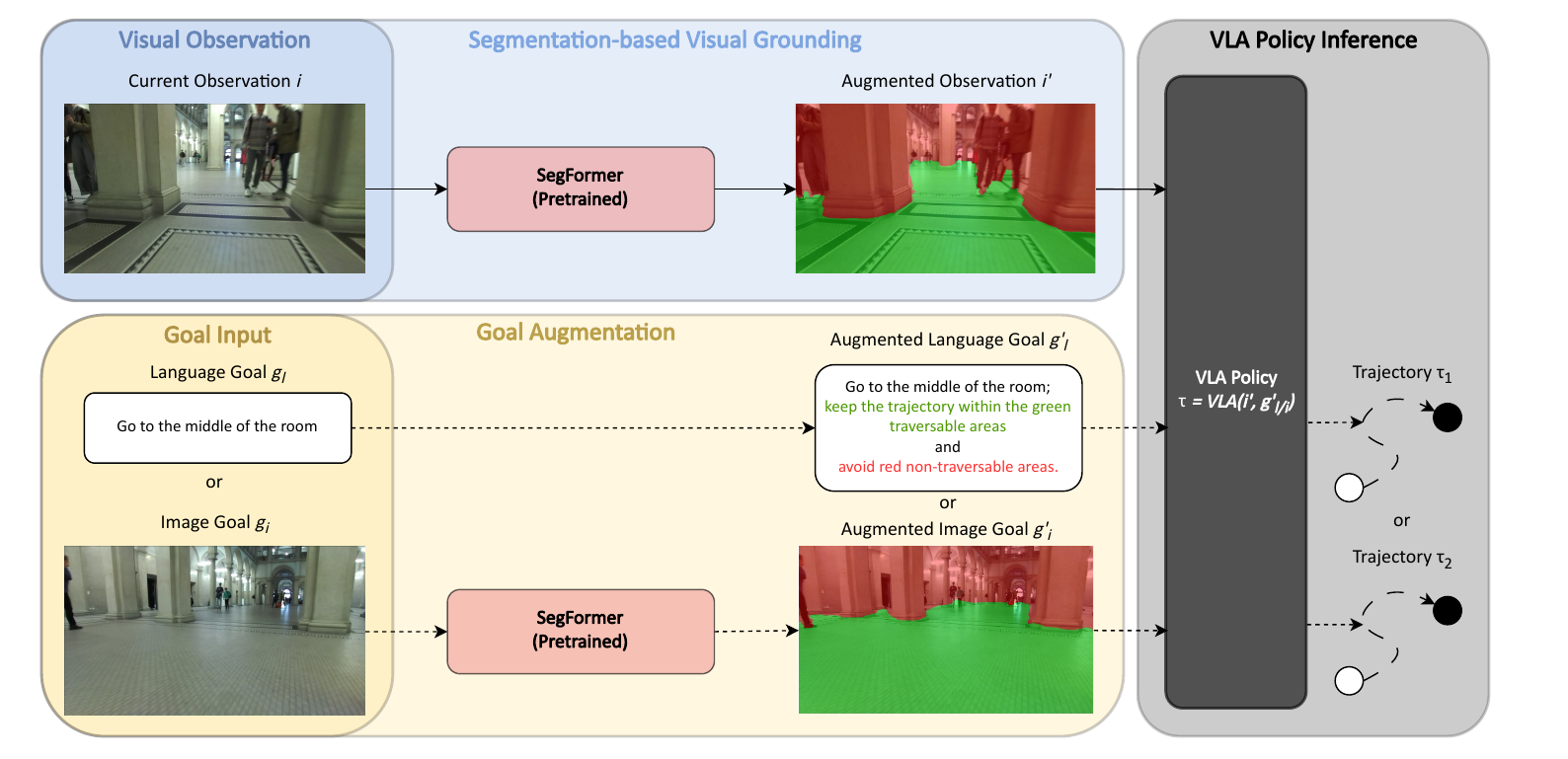}
    \caption{Overview of the proposed segmentation-based visual grounding and goal augmentation method. SegFormer produces real-time semantic segmentation to identify traversable (green) and non-traversable (red) regions. These overlays are used to augment both current observation and goals, which are then passed to the VLA policy to produce trajectories $\tau_1$ and $\tau_2$.}
    \label{fig:method_fig}
\end{figure*}

\section{Related work}\label{sec:related-work}

\subsection{Vision-Language and Vision-Language-Action Models for Navigation}

Large language models have seen increasing use, including for robot navigation and trajectory generation; because most robotic systems are equipped with cameras, vision-language models have emerged as a particularly promising approach for navigation. LM-NAV~\cite{shah2023lm} uses GPT-3~\cite{floridi2020gpt} to extract landmark names from a language navigation instruction, then uses CLIP~\cite{radford2021learning} to match those names to visual regions in the robot's camera image, enabling zero-shot outdoor navigation. VLMaps~\cite{huang2023visual} uses spatial map representation that directly fuses pretrained visual-language features with a 3D reconstruction of the physical world, allowing natural language-guided navigation. Recently, VL-TGS~\cite{song2025vl} generates multiple candidate trajectories using a CVAE-based approach \cite{abid2019contrastive} and selects the most suitable trajectory using a VLM with visual prompting.

As a more recent development, Vision-Language-Action models, while mostly used for manipulation tasks, have also been applied to navigation tasks.  MobilityVLA~\cite{chiang2024mobility} uses topological graphs to satisfy natural language or multimodal requests after observing a prerecorded demonstration tour. NaVILA~\cite{cheng2024navila} uses a 2-level framework that unifies a VLA policy with locomotion skills. Instead of directly predicting low-level actions from VLA, NaVILA first generates mid-level actions with spatial information in the form of a language instruction. OmniVLA~\cite{hirose2025omnivla} trains a VLA policy with three primary goal modalities: natural language, egocentric images, 2D poses, and their combinations. This allows the model to have a strong generalization to unseen environments and the ability to follow novel language instructions.

\subsection{Visual Grounding and Language Models}

Although large language models are great at generating sentences, their ability to reason, validate their outputs, and, in the case of vision language models, comprehend images remains limited \cite{huang2023language}. Therefore, visual grounding has been used to help language models understand and act on images more effectively. Visual grounding techniques help by preprocessing the image inputs by adding spatially relevant and easily comprehensible features. Set-of-Marks~\cite{yang2023set} allows GPT-4V~\cite{achiam2023gpt} to correctly answer questions about an image by segmenting using SEEM/SAM~\cite{zou2023segment, kirillov2023segment} and adding a numerical mark over each object, therefore helping the model to easily comprehend and refer to different objects. PIVOT~\cite{nasiriany2024pivot} casts robotic control as a VQA problem by annotating an image with a visual representation of robot actions and iteratively querying a VLM to select the most fitting action. BYOVLA~\cite{hancock2025run} uses GPT4-o~\cite{achiam2023gpt} to select task-irrelevant regions for manipulation tasks and fully mask irrelevant regions that distract the VLA, which allows the model to focus on the relevant parts without being distracted by backgrounds and irrelevant objects.

\section{Problem formulation}\label{sec:problem-formulation}

This paper presents an evaluation of a visual grounding technique in the context of a vision-language-action model used for robot navigation. The problem is formulated as a trajectory generation task predicted from the current visual observation and a goal given either by a language instruction or a visual goal in the form of an image.

The trajectory generation is defined as:
\begin{equation}
    \tau = VLA(i,g),
\end{equation}

where the input $i$ is the current egocentric observation image from the robot's camera, and $g \in \{g_l, g_i\}$ is either a language instruction $g_l$ or a goal image $g_i$. The output $\tau = \{(x_1, y_1), … (x_7, y_7)\}$ is a sequence of 7 waypoints in a 2D coordinate system that represents the predicted navigation path.

\section{Method}\label{sec:method}

We propose a visual grounding pipeline that overlays semantic segmentation onto egocentric observations before feeding them to a VLA policy. The grounding has two variants: observation-only grounding, which segments only the current visual observation, and joint grounding, which additionally augments the goal modality. The overview of the method can be seen in Figure~\ref{fig:method_fig}.

\subsection{Segmentation-Based Visual Grounding}

Given an egocentric RGB image from the robot's camera, we apply a semantic segmentation model to classify each pixel as either traversable or non-traversable. We use SegFormer~\cite{xie2021segformer}, a transformer-based segmentation framework with a hierarchical encoder and a lightweight MLP decoder, for binary segmentation, producing a pixel-wise mask where traversable regions (floor) are distinguished from non-traversable obstacles (walls, furniture, stairs).

The resulting mask is overlaid onto the original image: traversable areas are highlighted in green, non-traversable areas in red. This produces the grounded observation $i'$. The input segmentation is shown in Figure~\ref{fig:method_fig} as the blue box.

For observation-only grounding, the VLA generates a trajectory as follows:
\begin{equation}
    \tau_1 = VLA(i',g),
\end{equation}

For joint grounding, the goal is also augmented. For image goals, the same segmentation overlay is applied to the goal image. For language goals, we append an explicit specification to the instruction: ``keep the trajectory within green traversable areas and avoid red non-traversable areas''. This yields the augmented goal $g' \in \{g'_l, g'_i\}$. The goal augmentation is shown in Figure~\ref{fig:method_fig} as the yellow box.

The trajectory for the joint grounding, where the input is segmented and goal is augmented, is then:
\begin{equation}
    \tau_2 = VLA(i',g'),
\end{equation}

The design choices were constrained by three requirements: real-time inference, generalization to unseen environments (no fixed object vocabulary), and multimodal goal support (language and images).
LLM-based grounding methods such as Set-of-Marks~\cite{yang2023set} require chain-of-thought prompting, which increases the inference time from $0.5s$ to $5s$ per image, which is infeasible for real-time navigation. Object detection methods such as YOLO~\cite{terven2023comprehensive} are fast but rely on a fixed set of object classes, severely limiting generalization to novel obstacles in unseen environments. Segmentation-based grounding avoids both issues: SegFormer runs in real time and generalizes to any traversable/non-traversable distinction without a predefined vocabulary.

\section{Experiments}\label{sec:experiments}

This section covers the experimental setup, model and  dataset choice, what experiments were conducted, and the results from all the experiments.

\subsection{Experimental Setup}

OmniVLA~\cite{hirose2025omnivla} has been chosen as the evaluated VLA model due to its competitive success rate among other contemporary VLA models and due to its flexibility when it comes to the goal modality, allowing the evaluation of visual grounding not only in the context of vision language navigation, but also in the context of vision goal navigation. The model predicts a trajectory consisting of 7 waypoints. The experiments were carried out on an NVIDIA Jetson Thor.

The dataset used in the experimental evaluation is the ETH-2 episode from the Grand Tour dataset \cite{frey2026grandtour}, which is the largest legged robotics dataset collected using an ANYmal D quadruped robot with multimodal sensors that capture images, depth, proprioception, and more. Out of the dataset a specific episode has been chosen that was indoor and had a sizable number of describable objects for language instructions. The episode has been split into segments, where for each segment the model was given a specific language instruction or the final image of the segment in the case of goal image experiments. All segments and their language goals can be seen in Table~\ref{table:language-goals}.

The experiments included 3 full runs of the chosen episode, where 1) the default \textit{omnivla-base} model checkpoint has been used, 2) the finetuned \textit{omnivla-finetuned-cast} model checkpoint has been used, which has been finetuned with the CAST~\cite{glossop2025cast} dataset that aims to improve performance when using simpler, atomic instructions, and 3) the default \textit{omnivla-base} model checkpoint, but using image goals as input instead of language goals. For each full run, the model generated 3 trajectories: the original one that used the original current visual observation and the goal, labelled in plots as \textit{predicted\_base}, and then two trajectories which use visual grounding; \textit{predicted\_segmented}, which uses the segmented current visual observation and the original goal, and \textit{predicted\_segmented\_augmented}, which uses the segmented current visual observation and the augmented goal.

\subsection{Experimental Results}

We evaluated three prediction conditions across three model configurations: (1) base (original observation and goal), (2) segmented (grounded observation only), and (3) augmented (grounded observation + augmented goal). Performance was measured as the minimum Euclidean distance from each predicted waypoint to the ground-truth trajectory. Waypoints are indexed from 1 (nearest) to 7 (farthest). A small number of segments of the evaluated episode have been omitted in the final evaluations. Due to the limitation of the model to generate a trajectory only in 2-dimensional spaces and the nature of the segmentation technique used that tends to mark stairs as a non-traversable region, the beginning of the episode where the robot walks up the stairs has been omitted, together with all the segments where the robot stops, as explained later.

\begin{figure}
    \centering
    \includegraphics[width=\linewidth]{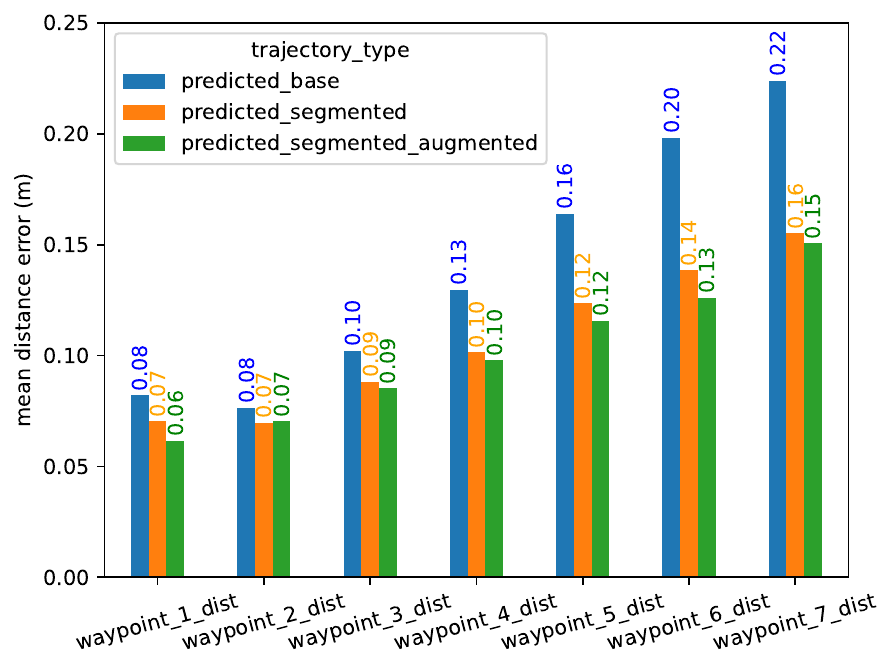}
    \caption{The mean error for each waypoint for the base prediction (model~=~\textit{omnivla-base}).}
    \label{fig:mean_distance_per_waypoint_omnivla-base}
\end{figure}

Visual grounding consistently reduced prediction error at all waypoints. For the \textit{omnivla-base} model with language goals, the average error at the farthest waypoint decreased from $0.22$~m (base) to $0.16$~m (segmented) and $0.15$~m (augmented); reductions of $27\%$ and $32\%$, respectively. The effect is more pronounced at farther waypoints; at waypoint 1, the reduction in error was only $13-25\%$. Figure~\ref{fig:mean_distance_per_waypoint_omnivla-base} visualizes this trend: the error-vs-distance curve is less steep for grounded predictions, indicating that grounding helps the most where the model otherwise struggles. The disaggregation by instruction length (Table~\ref{tab:table-errors-per-instruction-length}) shows that this far-waypoint benefit is greater for long instructions ($39-44\%$ error reduction at WP7) than for short instructions ($35-36\%$).

\begin{figure}
    \centering
    \includegraphics[width=\linewidth]{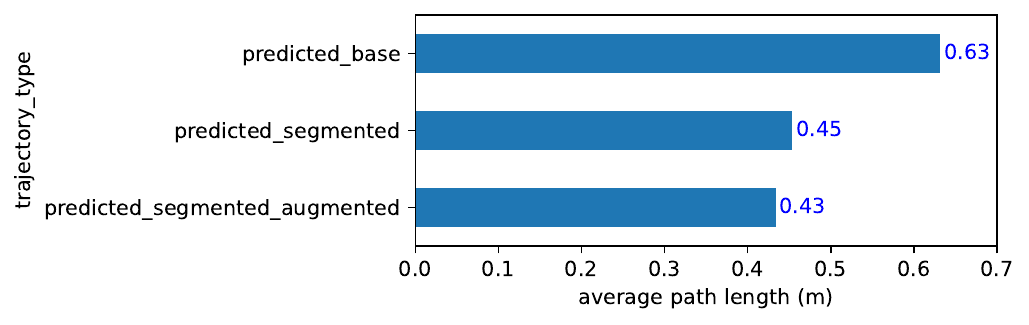}
    \caption{The average length of the predicted trajectories (model~=~\textit{omnivla-base}).}
    \label{fig:average_path_length_omnivla-base}
\end{figure}

\begin{figure}
    \centering
    \includegraphics[width=\linewidth]{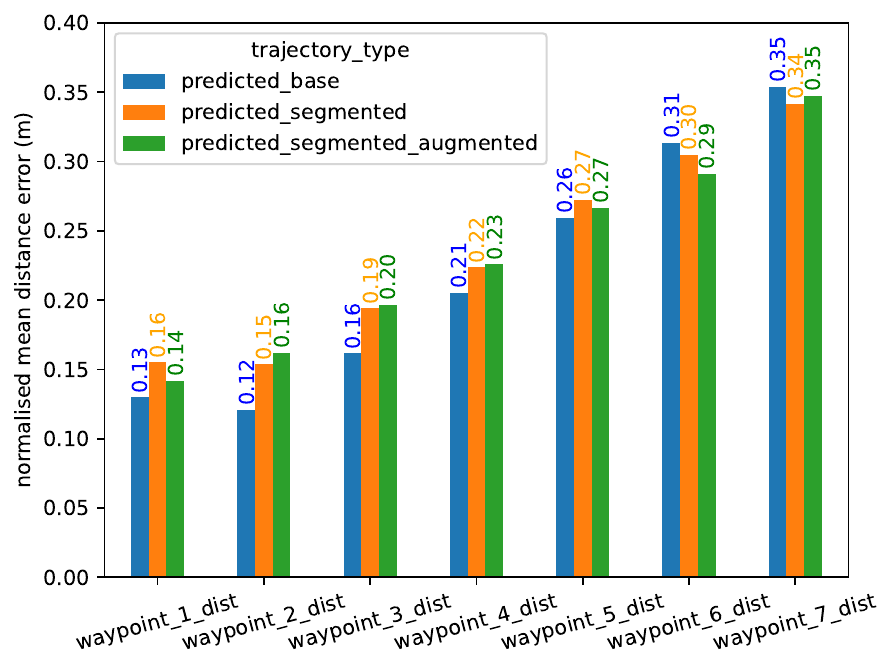}
    \caption{The normalised mean error for each waypoint for the base prediction (model~=~\textit{omnivla-base}).}
    \label{fig:normalised_mean_distance_per_waypoint_omnivla-base}
\end{figure}

However, grounded predictions also produced shorter trajectories. The base predictions averaged $0.63$~m in length, while the segmented and augmented predictions averaged $0.45$~m and $0.43$~m, respectively, as seen in Figure~\ref{fig:average_path_length_omnivla-base}. When we normalize error by trajectory length, the advantage of grounding largely disappears. As shown in Figure~\ref{fig:normalised_mean_distance_per_waypoint_omnivla-base}, normalized errors converge across all three conditions. In fact, at waypoints~$1-5$, base predictions show slightly lower normalized error than the grounded variants. This suggests that grounding primarily acts as a trajectory length regularizer rather than fundamentally improving the VLA's spatial reasoning per unit distance. 

\begin{figure}
    \centering
    \includegraphics[width=\linewidth]{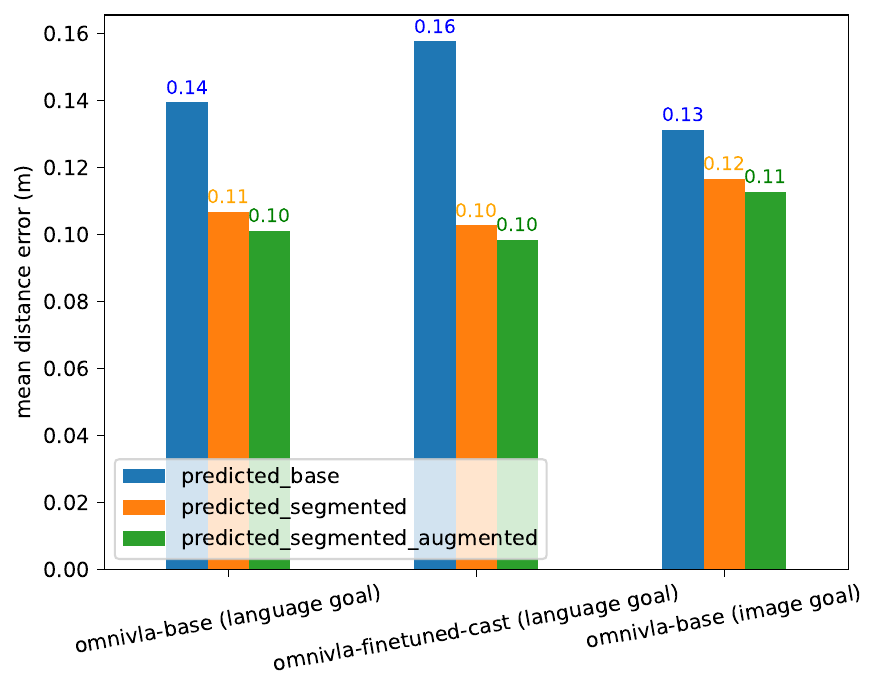}
    \caption{The overall mean distance error for each model/goal modality.}
    \label{fig:overall_mean_distance_error}
\end{figure}

We compared performance across three model runs: \textit{omnivla-base} with language goals, \textit{omnivla-finetuned-cast} (trained on atomic instructions), and \textit{omnivla-base} with image goals. The base + image configuration had the lowest base error of $0.13$~m, consistent with the reported strength of OmniVLA in visual goal navigation. Consequently, grounding provided little to no benefit and, in fact, slightly increased error for some segments. This indicates that visual grounding is more effective when paired with language goals than with image goals, likely because language instructions require the model to map abstract descriptions to spatial regions, a task that explicit traversability cues directly support. Figure~\ref{fig:overall_mean_distance_error} summarizes the results for each model/goal modality. The base + image configuration achieved $0.13$~m average error, and grounding provided a very small improvement $0.12-0.11$~m, compared to language goals where grounding reduced the error from $0.14-0.16$~m to $0.10-0.11$~m.

\begin{figure}
    \centering
    \includegraphics[width=\linewidth]{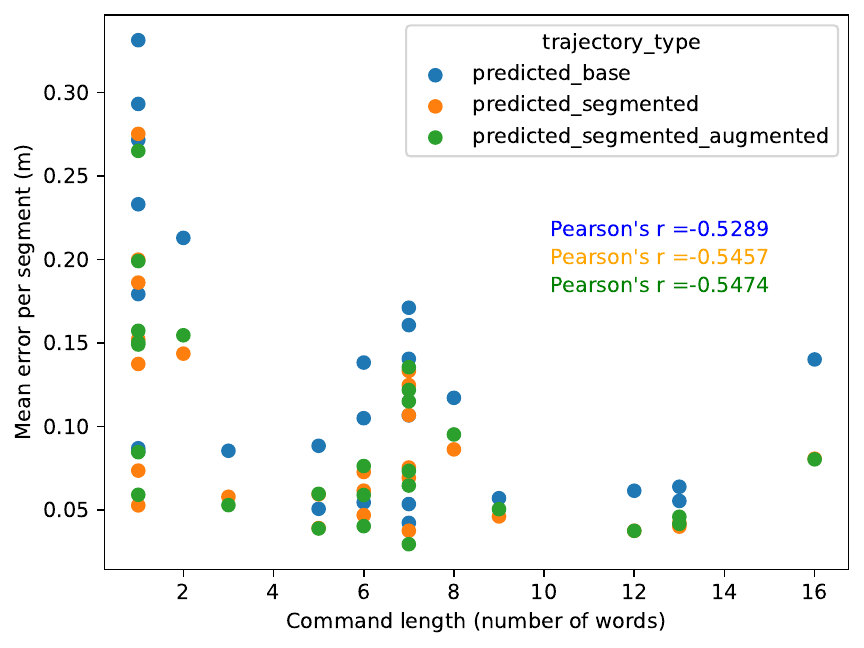}
    \caption{Mean error per segment vs instruction length of that segment (model~=~\textit{omnivla-finetuned-cast}).}
    \label{fig:mean_error_per_segment_vs_instruction_length_omnivla-finetuned-cast}
\end{figure}

We observed a mild negative correlation $r\approx-0.54$ between the number of words in the language instruction and the trajectory error (Figure~\ref{fig:mean_error_per_segment_vs_instruction_length_omnivla-finetuned-cast}). Instructions with 8 to 16 words produced a consistently lower error than single-word instructions. An extra outlier was the instruction “stop” (1 word), which the model failed to execute entirely.

\begin{table}[]
\textbf{}
    \begin{tabular}{|l|l|l|l|l|}
    \hline
    \textbf{Instruction length} & \multicolumn{1}{l|}{\textbf{Waypoint}} & \multicolumn{1}{l|}{\textbf{Base}} & \multicolumn{1}{l|}{\textbf{Segmented}} & \textbf{Augmented} \\ \hline
    "Stop" only                          & 1        & 0.204 & 0.149          & \textbf{0.100} \\ \cline{2-5} 
                                         & 1-7      & 0.274 & 0.174          & \textbf{0.161} \\ \hline
    Short                                & 1 (near) & 0.081 & 0.069          & \textbf{0.060} \\ \cline{2-5} 
    (2 - 7 words) & 3        & 0.053 & 0.048          & \textbf{0.047} \\ \cline{2-5} 
                                         & 5        & 0.108 & 0.074          & \textbf{0.071} \\ \cline{2-5} 
                                         & 7 (far)  & 0.187 & \textbf{0.120} & 0.122          \\ \hline
    Long                                 & 1        & 0.074 & 0.050          & \textbf{0.048} \\ \cline{2-5} 
    ( \textgreater{}= 8 words)           & 3        & 0.054 & \textbf{0.045} & \textbf{0.045} \\ \cline{2-5} 
                                         & 5        & 0.098 & \textbf{0.060} & 0.061          \\ \cline{2-5} 
                                         & 7        & 0.160 & \textbf{0.090} & 0.098          \\ \hline
    Overall average                      &          &       &                &                \\ 
    (without stop)                       & 1-7      & 0.095 & \textbf{0.064} & \textbf{0.064} \\ \hline
    \end{tabular}
    \caption{Per-waypoint prediction error (meters) by grounding type, disaggregated by instruction length. (Values show mean Euclidean distance from predicted to ground-truth waypoints. Short instructions ($13$ segments) include atomic instructions such as ``turn left''; long instructions ($6$ segments) are descriptive navigation goals. Best-performing grounding condition per row is bolded.)}
    \label{tab:table-errors-per-instruction-length}
\end{table}

The model never generated a zero-displacement trajectory when given the language goal “stop” or an image goal identical to the current observation. Instead, it consistently predicted forward motion. We hypothesize that the VLA's training data lacked stationary instructions, making this an out-of-distribution input. This failure persists across all three model variants and both grounding conditions, indicating that perceptual augmentation alone cannot compensate for the missing training signal. As quantified in the ``stop''-only row of Table~\ref{tab:table-errors-per-instruction-length}, the model's error at WP1 ranges from $0.1$~m (augmented) to $0.204$~m (base) instead of the ideal $0$, with the average showing persistent error between $0.161$~m and $0.274$~m across all grounding conditions.

\begin{figure}
    \centering
    \includegraphics[width=\linewidth]{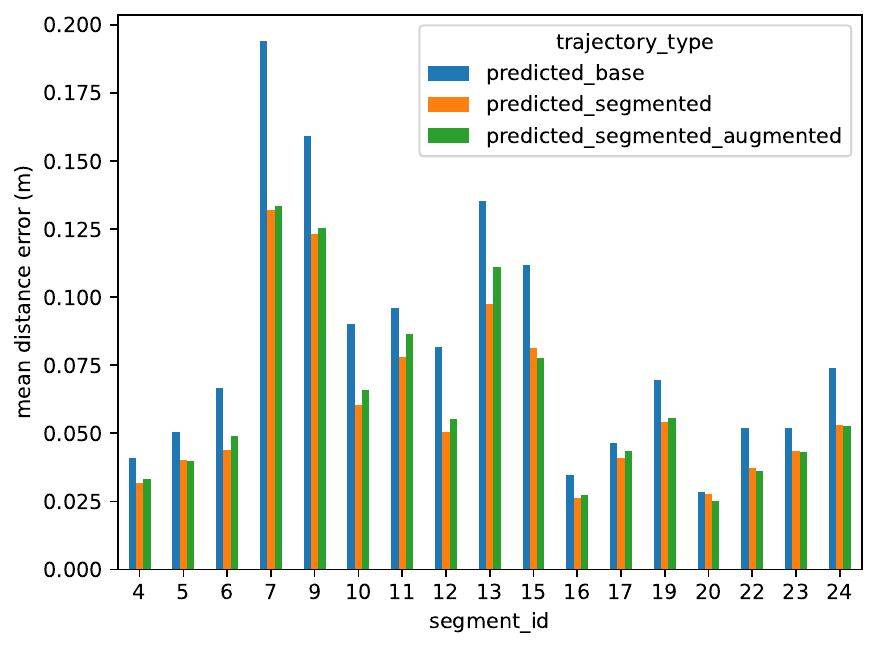}
    \caption{The mean error for each segment by trajectory type, stops and stair segments removed (model~=~\textit{omnivla-base}).}
    \label{fig:error_per_segment_by_trajectory_type_without_stops_omnivla-base}
\end{figure}

For approximately half of the segments, joint grounding (augmented goals) did not perform better than observation-only grounding (Figure~\ref{fig:error_per_segment_by_trajectory_type_without_stops_omnivla-base}). Augmentation helped most for ambiguous instructions, but offered little advantage when instructions were already descriptive. This suggests that the appended instruction is redundant when the language goal already implies traversability constraints. See Table~\ref{table:language-goals} in the Appendix for all language instructions.

\section{Limitations}\label{sec:limitations}

Despite promising results, our approach has inherent limitations. First, the segmentation method assumes static, 2D traversability, therefore dynamic obstacles and 3D structures (e.g. stairs) are not handled. Second, the red/green overlay is a simple heuristic; optimal color mapping or segmentation encoding for VLA inputs remain unexplored. Third, our grounding requires the VLA to interpret visual overlays correctly, and there is no guarantee that other VLAs with different vision encoders would respond similarly. Fourth, the ``stop'' instruction failure is not solved by grounding, indicating that certain atomic instructions may require explicit training data rather than perceptual augmentation.

Ultimately, the results of this preliminary study are limited to a single indoor episode and should be interpreted with caution.

\section{Conclusion}\label{sec:conclusion}

To our knowledge, this paper presents the first empirical evaluation of visual grounding for vision-language-action navigation policies. We introduced a real-time, segmentation-based grounding method using SegFormer that satisfies the three constraints for navigation: real-time inference, generalization to unseen environments, and multimodal goal support. Through ablative analysis on the Grand Tour dataset using OmniVLA, we demonstrated that visual grounding consistently reduced waypoint prediction error, particularly at the farthest waypoint ($27-44\%$ reduction depending on instruction length), and is substantially more effective for language goals than for image goals.

Our results reveal several important insights. First, while grounding reduces the absolute far-waypoint error by $27-44\%$, normalized error becomes comparable to baseline, and grounded trajectories are $30\%$ shorter, confirming that grounding primarily acts as a trajectory length regularizer rather than fundamentally improving spatial reasoning. Second, the mild negative correlation between instruction verbosity and error indicates that VLA models benefit from more descriptive language goals. Third, the consistent failure to generate zero-displacement ``stop'' trajectories reveals a critical gap in current VLA training paradigms, as the model cannot represent or execute stationary instructions even when explicitly instructed.

Within the boundaries of our limitations, we have shown that visual grounding offers a simple, computationally inexpensive, and effective method to improve VLA navigation performance without model retraining.

Future work can extend this approach in several directions. Online closed-loop evaluation would validate whether the observed open-loop error reductions translate to improved task completion rates. Extending grounding to handle dynamic obstacles and 3D structures (e.g. stairs) would address the most significant limitation of the current method. Finally, the ``stop'' instruction failure suggests that combining perceptual grounding with targeted fine-tuning on atomic instructions may be necessary for full coverage of primitive navigation commands.

\section*{Acknowledgments}

This work was supported by the CDT in Cybersecurity [EP/S022503/1] and the UKRI FLF [MR/V025333/1] (RoboHike). For the purpose of Open Access, the author has applied a CC BY public copyright license to any Author Accepted Manuscript version arising from this submission.

\bibliographystyle{plainnat}
\bibliography{references}

@article{hirose2025omnivla,
  title={OmniVLA: An omni-modal vision-language-action model for robot navigation},
  author={Hirose, Noriaki and Glossop, Catherine and Shah, Dhruv and Levine, Sergey},
  journal={arXiv preprint arXiv:2509.19480},
  year={2025}
}

@article{cheng2024navila,
  title={Navila: Legged robot vision-language-action model for navigation},
  author={Cheng, An-Chieh and Ji, Yandong and Yang, Zhaojing and Gongye, Zaitian and Zou, Xueyan and Kautz, Jan and B{\i}y{\i}k, Erdem and Yin, Hongxu and Liu, Sifei and Wang, Xiaolong},
  journal={arXiv preprint arXiv:2412.04453},
  year={2024}
}

@article{chiang2024mobility,
  title={Mobility vla: Multimodal instruction navigation with long-context vlms and topological graphs},
  author={Chiang, Hao-Tien Lewis and Xu, Zhuo and Fu, Zipeng and Jacob, Mithun George and Zhang, Tingnan and Lee, Tsang-Wei Edward and Yu, Wenhao and Schenck, Connor and Rendleman, David and Shah, Dhruv and others},
  journal={arXiv preprint arXiv:2407.07775},
  year={2024}
}

@article{yang2023set,
  title={Set-of-mark prompting unleashes extraordinary visual grounding in gpt-4v},
  author={Yang, Jianwei and Zhang, Hao and Li, Feng and Zou, Xueyan and Li, Chunyuan and Gao, Jianfeng},
  journal={arXiv preprint arXiv:2310.11441},
  year={2023}
}

@article{nasiriany2024pivot,
  title={Pivot: Iterative visual prompting elicits actionable knowledge for vlms},
  author={Nasiriany, Soroush and Xia, Fei and Yu, Wenhao and Xiao, Ted and Liang, Jacky and Dasgupta, Ishita and Xie, Annie and Driess, Danny and Wahid, Ayzaan and Xu, Zhuo and others},
  journal={arXiv preprint arXiv:2402.07872},
  year={2024}
}

@inproceedings{shah2023lm,
  title={Lm-nav: Robotic navigation with large pre-trained models of language, vision, and action},
  author={Shah, Dhruv and Osi{\'n}ski, B{\l}a{\.z}ej and Levine, Sergey and others},
  booktitle={Conference on robot learning},
  pages={492--504},
  year={2023},
  organization={pmlr}
}

@article{floridi2020gpt,
  title={GPT-3: Its nature, scope, limits, and consequences},
  author={Floridi, Luciano and Chiriatti, Massimo},
  journal={Minds and machines},
  volume={30},
  number={4},
  pages={681--694},
  year={2020},
  publisher={Springer}
}

@inproceedings{radford2021learning,
  title={Learning transferable visual models from natural language supervision},
  author={Radford, Alec and Kim, Jong Wook and Hallacy, Chris and Ramesh, Aditya and Goh, Gabriel and Agarwal, Sandhini and Sastry, Girish and Askell, Amanda and Mishkin, Pamela and Clark, Jack and others},
  booktitle={International conference on machine learning},
  pages={8748--8763},
  year={2021},
  organization={PmLR}
}

@inproceedings{huang2023visual,
  title={Visual language maps for robot navigation},
  author={Huang, Chenguang and Mees, Oier and Zeng, Andy and Burgard, Wolfram},
  booktitle={2023 IEEE International Conference on Robotics and Automation (ICRA)},
  pages={10608--10615},
  year={2023},
  organization={IEEE}
}

@article{song2025vl,
  title={Vl-tgs: Trajectory generation and selection using vision language models in mapless outdoor environments},
  author={Song, Daeun and Liang, Jing and Xiao, Xuesu and Manocha, Dinesh},
  journal={IEEE Robotics and Automation Letters},
  year={2025},
  publisher={IEEE}
}

@article{abid2019contrastive,
  title={Contrastive variational autoencoder enhances salient features},
  author={Abid, Abubakar and Zou, James},
  journal={arXiv preprint arXiv:1902.04601},
  year={2019}
}

@article{huang2023language,
  title={Language is not all you need: Aligning perception with language models},
  author={Huang, Shaohan and Dong, Li and Wang, Wenhui and Hao, Yaru and Singhal, Saksham and Ma, Shuming and Lv, Tengchao and Cui, Lei and Mohammed, Owais Khan and Patra, Barun and others},
  journal={Advances in Neural Information Processing Systems},
  volume={36},
  pages={72096--72109},
  year={2023}
}

@article{achiam2023gpt,
  title={Gpt-4 technical report},
  author={Achiam, Josh and Adler, Steven and Agarwal, Sandhini and Ahmad, Lama and Akkaya, Ilge and Aleman, Florencia Leoni and Almeida, Diogo and Altenschmidt, Janko and Altman, Sam and Anadkat, Shyamal and others},
  journal={arXiv preprint arXiv:2303.08774},
  year={2023}
}

@article{zou2023segment,
  title={Segment everything everywhere all at once},
  author={Zou, Xueyan and Yang, Jianwei and Zhang, Hao and Li, Feng and Li, Linjie and Wang, Jianfeng and Wang, Lijuan and Gao, Jianfeng and Lee, Yong Jae},
  journal={Advances in neural information processing systems},
  volume={36},
  pages={19769--19782},
  year={2023}
}

@inproceedings{kirillov2023segment,
  title={Segment anything},
  author={Kirillov, Alexander and Mintun, Eric and Ravi, Nikhila and Mao, Hanzi and Rolland, Chloe and Gustafson, Laura and Xiao, Tete and Whitehead, Spencer and Berg, Alexander C and Lo, Wan-Yen and others},
  booktitle={Proceedings of the IEEE/CVF international conference on computer vision},
  pages={4015--4026},
  year={2023}
}

@inproceedings{hancock2025run,
  title={Run-time observation interventions make vision-language-action models more visually robust},
  author={Hancock, Asher J and Ren, Allen Z and Majumdar, Anirudha},
  booktitle={2025 IEEE International Conference on Robotics and Automation (ICRA)},
  pages={9499--9506},
  year={2025},
  organization={IEEE}
}

@article{xie2021segformer,
  title={SegFormer: Simple and efficient design for semantic segmentation with transformers},
  author={Xie, Enze and Wang, Wenhai and Yu, Zhiding and Anandkumar, Anima and Alvarez, Jose M and Luo, Ping},
  journal={Advances in neural information processing systems},
  volume={34},
  pages={12077--12090},
  year={2021}
}

@article{terven2023comprehensive,
  title={A comprehensive review of yolo architectures in computer vision: From yolov1 to yolov8 and yolo-nas},
  author={Terven, Juan and C{\'o}rdova-Esparza, Diana-Margarita and Romero-Gonz{\'a}lez, Julio-Alejandro},
  journal={Machine learning and knowledge extraction},
  volume={5},
  number={4},
  pages={1680--1716},
  year={2023},
  publisher={mdpi}
}

@article{frey2026grandtour,
  title={Grandtour: A legged robotics dataset in the wild for multi-modal perception and state estimation},
  author={Frey, Jonas and Tuna, Turcan and Fu, Frank and Patterson, Katharine and Xu, Tianao and Fallon, Maurice and Cadena, Cesar and Hutter, Marco},
  journal={arXiv preprint arXiv:2602.18164},
  year={2026}
}

@article{glossop2025cast,
  title={Cast: Counterfactual labels improve instruction following in vision-language-action models},
  author={Glossop, Catherine and Chen, William and Bhorkar, Arjun and Shah, Dhruv and Levine, Sergey},
  journal={arXiv preprint arXiv:2508.13446},
  year={2025}
}

\appendix

\begin{table}[h]
\begin{tabular}{lll}
segment\_id & start\_s & language\_goal                                 \\ \hline
0           & 0        & Stop                                           \\
1           & 6        & Go up the middle of the stairs                 \\
2           & 16       & Stop                                           \\
3           & 33       & Go up the middle of the stairs                 \\
4           & 37       & Go forward through the middle archway          \\
5  & 54  & \begin{tabular}[c]{@{}l@{}}Turn left after passing the left pillar and go towards\\ the gray cabinet\end{tabular}           \\
6           & 74       & Keep going straight down the corridor          \\
7           & 88       & Turn right                                     \\
8           & 93       & Stop                                           \\
9           & 124      & Turn left; go forward; and turn right          \\
10          & 130      & Go forward and towards the right pillar        \\
11          & 139      & Go between the two pillars; then turn right    \\
12          & 149      & Go forward past the white boxes                \\
13          & 165      & Turn right 45 degrees and go forward           \\
14          & 171      & Stop                                           \\
15 & 182 & \begin{tabular}[c]{@{}l@{}}Turn right and go straight down the hallway and\\  to the right of the right pillar\end{tabular} \\
16          & 221      & Go forward between the pillars                 \\
17          & 236      & Go forward and around the fountain on the left \\
18          & 256      & Stop                                           \\
19          & 262      & Go forward between the pillar                  \\
20          & 283      & Go to the middle of the room                   \\
21          & 291      & Stop                                           \\
22 & 299 & Go to the other side of the room along the right pillars                                                                    \\
23 & 322 & \begin{tabular}[c]{@{}l@{}}Go to the other side of the room and through\\ the middle archway\end{tabular}                   \\
24          & 338      & Keep going forward                             \\
25          & 346      & Stop                                          
\end{tabular}
\caption{Table of all the language goals for each segment and their corresponding segment id and starting second in the Grand Tour episode ETH-2.}
\label{table:language-goals}
\end{table}

\end{document}